\documentclass[final]{cvpr}

\usepackage{graphicx}

\usepackage{times}
\usepackage{epsfig}
\usepackage{graphicx}
\usepackage{amsmath}
\usepackage{amssymb}
\usepackage{multirow}
\usepackage[numbers]{natbib}
\usepackage{algorithmic}
\usepackage{algorithm}

\newcommand{\cutout}{\texttt{Cutout}~}
\newcommand{\cutmix}{\texttt{CutMix}~}
\newcommand{\randaug}{\texttt{RandAugment}~}
\newcommand{\autoaug}{\texttt{AutoAugment}~}

\usepackage[pagebackref=true,breaklinks=true,colorlinks,bookmarks=false]{hyperref}

\def\cvprPaperID{7376} 
\def\confYear{CVPR 2021}

\newcommand{\ind}{\mathbb{I}}

\newcommand{\I}{{\mathcal{I}}}

\begin{document}

\title{KeepAugment: A Simple Information-Preserving\\ Data Augmentation Approach}

\author{
Chengyue Gong\textsuperscript{1}, Dilin Wang\textsuperscript{2}, Meng Li\textsuperscript{2}, Vikas Chandra\textsuperscript{2}, Qiang Liu\textsuperscript{1} \\
\\
\textsuperscript{1} University of Texas, Austin \hspace{20pt} \textsuperscript{2} Facebook Inc
}

\maketitle

\begin{abstract}
Data augmentation (DA) is an essential technique for training state-of-the-art deep learning systems. In this paper, we empirically show data augmentation might introduce noisy augmented examples and consequently hurt the performance on unaugmented data during inference. To alleviate this issue, we propose a simple yet highly effective approach, dubbed \emph{KeepAugment}, to increase augmented images fidelity. The idea is first to use the saliency map to detect important regions on the original images and then preserve these informative regions during augmentation. This information-preserving strategy allows us to generate more faithful training examples. Empirically, we demonstrate our method significantly improves on a number of prior art data augmentation schemes, e.g. AutoAugment, Cutout, random erasing, achieving promising results on image classification, semi-supervised image classification, multi-view multi-camera tracking and object detection. 
\end{abstract}

\section{Introduction}

\begin{figure*}[t]
\centering
\setlength{\tabcolsep}{1pt}
\begin{tabular}{cccccc}
\includegraphics[height=0.15\textwidth]{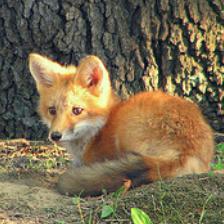} &
\includegraphics[height=0.15\textwidth]{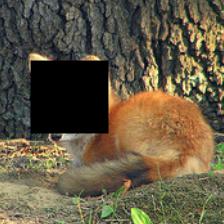} & 
\includegraphics[height=0.15\textwidth]{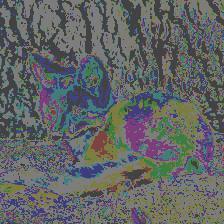}  ~~~&
\includegraphics[height=0.15\textwidth]{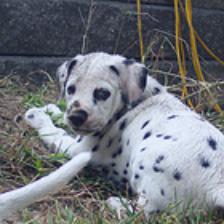} &
\includegraphics[height=0.15\textwidth]{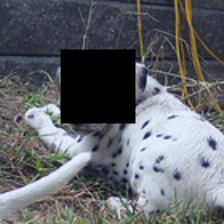} & 
\includegraphics[height=0.15\textwidth]{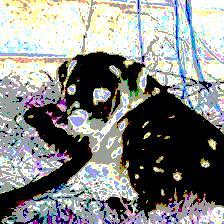}  \\
{\scriptsize (a1) Red fox} & {\scriptsize (a2) Cutout} & {\scriptsize (a3) RandAugment } & {\scriptsize (b1) Dog } & {\scriptsize (b2) Cutout} & {\scriptsize (b3) RandAugment } \\ 
\includegraphics[height=0.15\textwidth]{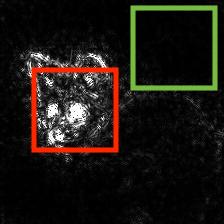} &
\includegraphics[height=0.15\textwidth]{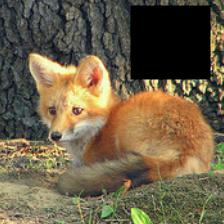} &
\includegraphics[height=0.15\textwidth]{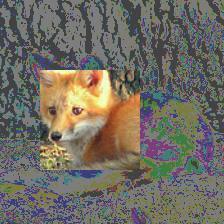} ~~~&
\includegraphics[height=0.15\textwidth]{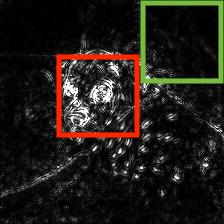}&
\includegraphics[height=0.15\textwidth]{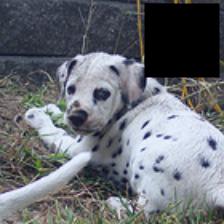} &
\includegraphics[height=0.15\textwidth]{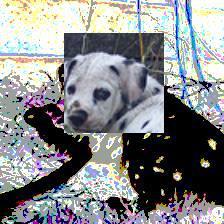} \\
{\scriptsize (a4) Saliency map} & {\scriptsize (a5) Keep+Cutout} & {\scriptsize (a6) Keep+RandAugment} & {\scriptsize (b4) Saliency map} & {\scriptsize (b5) Keep+Cutout} & {\scriptsize (b6) Keep+RandAugment} \\ 
\end{tabular}
\caption{
KeepAugment improves existing data augmentation by
always keeping the important regions (measured using saliency map) of the image untouched during augmentation. 
This is achieved by either avoiding to cut important regions (see KeepCutout), or pasting important regions on top of the transformed images (see KeepRandAugment). Images are from ImageNet \citep{deng2009imagenet}.
}
\label{fig:demo}
\end{figure*}

Recently, data augmentation is proving to be a crucial technique for solving
various challenging deep learning tasks, including image classification \citep[e.g.][]{devries2017cutout, yun2019cutmix, cubuk2018autoaugment, cubuk2019randaugment}, natural language understanding \citep[e.g.][]{devlin2018bert}, speech recognition \citep{park2020improved} and semi-supervised learning \citep[e.g.][]{xie2019unsupervised, sohn2020fixmatch, berthelot2019mixmatch}. 
Notable examples include regional-level augmentation methods, such as Cutout ~\citep{devries2017cutout} and {CutMix} \citep{yun2019cutmix}, which mask or modify randomly selected rectangular regions of the images; and 
image-level augmentation approaches, such as AutoAugment \citep{cubuk2018autoaugment} and Fast Augmentation \citep{lim2019fast}), 
which leverage reinforcement learning
to find optimal policies for selecting and combining different label-invariant transforms (e.g., rotation, color-inverting, flipping).
Although data augmentation increases the effective data size and 
promotes diversity in training examples, 
it inevitably introduces 
noise and ambiguity 
into the training process. Hence the overall performance would deteriorate if the augmentation is not properly modulated. 
For example,  
as shown in Figure~\ref{fig:demo}, 
random \cutout~(Figure~\ref{fig:demo} (a2) and (b2)) or \randaug~(Figure~\ref{fig:demo} (a3) and (b3)) 
may destroy the key characteristic information of original images 
that is responsible for classification, 
creating augmented images to have wrong or ambiguous labels. 


In this work, we propose \emph{KeepAugment}, 
a simple yet powerful 
adaptive data augmentation 
approach that aims to increase the fidelity of data  augmentation 
by \emph{always keeping important regions untouched} during augmentation. 
The idea is very simple: at each training step, we first score the importance of different regions of the original images using attribution methods such as saliency-map \citep{simonyan2013deep}; 
then we perform data augmentation in an adaptive way, such that regions with high importance scores always remain intact.  
This is achieved by either 
avoiding cutting critical high-score areas (see Figure~\ref{fig:demo}(a5) and  (b5)), or pasting the patches with high importance scores to the augmented images
(see Figure~\ref{fig:demo}(a6) and (b6)). 

Although KeepAugment is very simple and not resource-consuming,  
the empirical results on a variety of vision tasks show that we can significantly improve prior art data augmentation (DA) baselines. 
Specifically, 
for image classification, 
we achieve substantial improvements on existing DA techniques, including \texttt{Cutout} \citep{devries2017cutout}, \texttt{AutoAugment} \citep{cubuk2018autoaugment}, and \texttt{CutMix} \citep{yun2019cutmix}, 
boosting the performance on CIFAR-10 and ImageNet across various neural architectures.
In particular, we achieve $98.7\%$ test accuracy on CIFAR-10 using PyramidNet-ShakeDrop \citep{yamada2018shakedrop} by applying our method on top of \texttt{AutoAugment}.
When applied to multi-view multi-camera tracking, 
we improve upon the recent state-of-the-art results
on the Market1501 \citep{zheng2015market1501} dataset. 
In addition, we demonstrate that our method can be applied to semi-supervised learning and 
the model trained on ImageNet using our method 
can be transferred to COCO 2017 objective detection tasks \citep{lin2014microsoft} and 
allows us to improve the strong \texttt{Detectron2} baselines \citep{wu2019detectron2}.  


\section{Data Augmentation}
In this work, we focus on label-invariant data augmentation due to their popularity 
and significance in boosting empirical performance in practice.  
Let $x$ be an input image, data augmentation techniques allow us  to generate new images $x' = \mathcal A(x)$ that are expected to  have the same label as $x$, where $\mathcal A$ denotes a  label-invariant image transform, which is typically a stochastic function. 
Two classes of augmentation techniques are widely used for achieving state-of-the-art results on computer vision tasks: 

\paragraph{Region-Level Augmentation}
Region-level augmentation schemes, 
including Cutout \citep{devries2017cutout} and random erasing \citep{zhong2017random}, 
work by randomly masking out or modifying rectangular  regions of the input images, 
thus creating partially occluded data examples outside the span of the training data. 
This procedure could be conveniently formulated as applying randomly generated binary masks to 
the original inputs.  Precisely, consider an input  image $x$ of size $H  \times W$, 
and a rectangular  region  $S$ of the image domain.
Let  $M(S) = [M_{ij}(S)]_{ij}$ be the binary mask of $S$ with $M_{ij}(S) = \ind((i,j)\in S)$.
Then the augmented data can be generated by modifying the image on region $S$, yielding images of form 
${x'} = (1-M(S)) \odot x + M(S)\odot \delta$, 
where 
$\odot$ is element-wise multiplication, and 
 $\delta$ can be either zeros (for Cutout) or 
 random numbers (for random erasing). 
See Figure~\ref{fig:demo}(a2) and (b2) for examples. 

\paragraph{
Image-Level Augmentation} 
Exploiting  the invariance properties of natural images, 
image-level augmentation methods  
apply label-invariant transformations on the whole  image, such as solarization, sharpness, posterization, and color normalization.  
Traditionally, 
image-level transformations are often manually designed and heuristically chosen. 
Recently, AutoAugment~\citep{cubuk2018autoaugment} applies reinforcement learning to automatically  search optimal compositions of transformations.  
Several subsequent works, including RandAugment  \citep{cubuk2019randaugment}, Fast AutoAugment \citep{lim2019fast}, 
alleviate the heavy computational burden of searching on the space of transformation policies by
designing  more compact search spaces.  See Figure ~\ref{fig:demo}(b3) and Figure~\ref{fig:demo}(a3) for examples of transforms used by RandAugment.

\begin{figure*}[t]
\centering
\setlength{\tabcolsep}{1pt}
\begin{tabular}{ccc}
\raisebox{1em}{\rotatebox{90}{ Cifar-10 Accuracy}}
\includegraphics[height=0.2\textwidth]{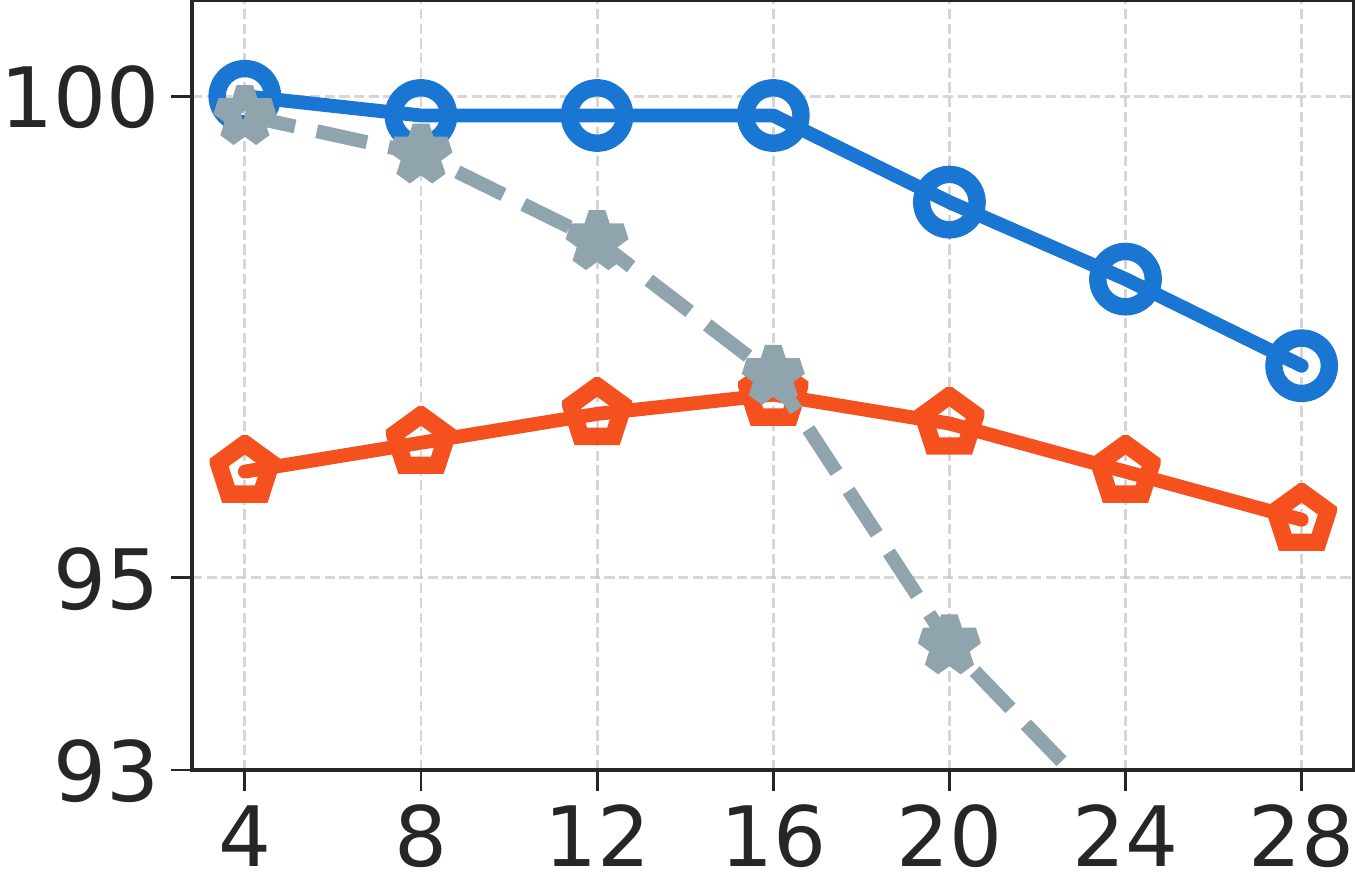} &
\raisebox{1em}{\rotatebox{90}{ Cifar-10 Accuracy}}
\includegraphics[height=0.2\textwidth]{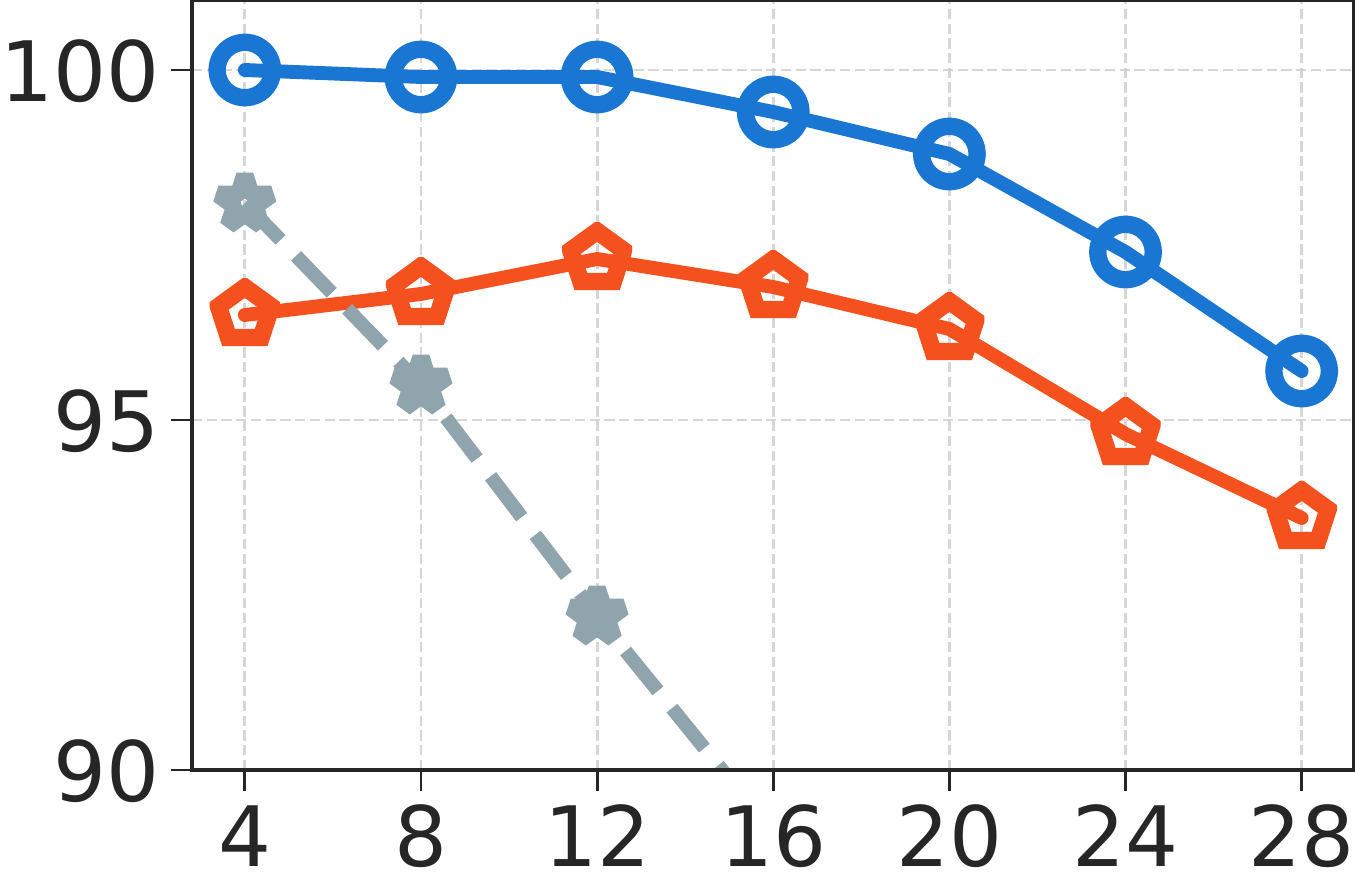} &
\hspace{0.5em}
\raisebox{1.5em}{\includegraphics[height=0.08 \textwidth]{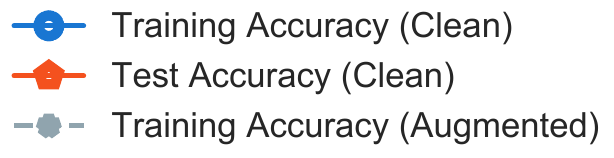}} \\
 {\small (a) Cutout length in CutOut} & {\small(b) Distortion magnitude in RandAugment}  \\
\end{tabular}
\caption{
The training and testing accuracy 
of Wide ResNet-28-10 trained 
on CIFAR-10 with Cutout and RandAugment, 
when we 
vary the cutout length of Cutout (a), 
and the distortion magnitude of RandAugment (b). 
We follow the same implementation details as in \citep{devries2017cutout} and \citep{cubuk2019randaugment}. 
For RandAugment, 
we fix the number of transformations to be 3 as  suggested in \citep{cubuk2019randaugment}.  
}
\label{fig:stronger_data_augmentation}
\end{figure*}

\vspace{-10pt}
\paragraph{Data Augmentation and its Trade-offs}
Although data augmentation increases the effective  size of data, 
it may inevitably cause loss of information and introduce noise and ambiguity  
if the augmentation is not controlled properly \citep[e.g.][]{wei2020circumventing, gontijo2020affinity}.  
To study this phenomenon empirically,  
we plot the train and testing accuracy on CIFAR-10  \citep{krizhevsky2009learning} when we apply  {\cutout} with increasingly large cutout length
in Figure~\ref{fig:stronger_data_augmentation}(a), 
and RandAugment with increasing distortion magnitude (see \citep{cubuk2019randaugment} for the definition) in Figure~\ref{fig:stronger_data_augmentation}(b). 
As typically expected, 
the generalization (the gap between the training and testing accuracy on clean data)
improves as the magnitude of the transform 
increases in both cases. 
However, when the magnitudes of the transform are too large ($\geq 16$ for {\cutout} and $\geq 12$ for {\randaug}), the training accuracy (blue line), 
and hence the testing accuracy (red line), 
starts to degenerate, 
indicating that augmented data no longer faithfully represent the clean training data in this case, 
such that the training loss on augmented data 
no longer forms a good surrogate of the training loss on the clean data.  

\section{Our Method}\label{sec:mainmethod}
We introduce our method for 
controlling the fidelity of data augmentation
and hence decreasing harmful misinformation. 
Our idea is to measure the importance of  the rectangular regions 
in the image by saliency map, 
and ensure that the regions with the highest scores are always presented after the data augmentation: 
for \cutout, we achieve this by avoiding to cut the important regions (see Figure~\ref{fig:demo}(a5) and (b5)); 
for image-level transforms such as RandAugment, 
we achieve this by \emph{pasting} the important regions on the top of the transformed images (see Figure~\ref{fig:demo}~ (a6) and (b6)).  

Specifically, let $g_{ij}(x, y)$ be saliency map of an image $x$ 
on pixel $(i,j)$ with the given label $y$.  
For a region $S$ on the image, 
its importance score is defined by 
\begin{align}
\mathcal I(S,x,y) = \sum_{(ij)\in S}g_{ij}(x, y).
\label{eq:saliency}
\end{align}
In our work, we use the standard saliency map based on vanilla gradient \citep{simonyan2013deep}. 
Specifically, given an image $x$ and its corresponding label logit value $\ell_y(x)$, 
we take $g_{ij}(x, y)$ to be the absolute value of vanilla gradients $|\nabla_x \ell_y(x)|$. For RBG-images, 
we take channel-wise maximum to get a single saliency value for each pixel $(i,j)$.

\paragraph{Selective-Cut}
For region-level (e.g. cutout-based) augmentation that masks or modifies randomly selected rectangle regions, 
we control the fidelity of data augmentation by ensuring that the regions being cut can not have large importance scores. 
This is achieved in practice by Algorithm~\ref{alg:main}(a), 
in which we randomly sample regions $S$ to be cut 
until its importance score $\I(S, x, y)$  is smaller than a given threshold $\tau$. 
The corresponding augmented example is defined as follows,
\begin{align}
    \Tilde{x} = (1-M(S)) \odot x,
    \label{eq:selective_cut}
\end{align}
where $M(S) = [M_{ij}(S)]_{ij}$ is the binary mask for $S$, 
with $M_{ij} = \ind((i,j)\in S)$.  
%

\paragraph{Selective-Paste} 
Because image-level transforms modify the whole images jointly, 
we ensure the fidelity of the transform by 
pasting a random region with high importance score (see Figure \ref{fig:demo}(a6) and (b6) for an example).
Algorithm~\ref{alg:main}(b) shows how we achieve this in practice, 
in which we draw an image-level augmented data $x' = \mathcal A(x)$,  
uniformly sample a region $S$ that satisfies $\mathcal I(S, x, y) > \tau $ for a threshold $\tau$, and  and paste the region $S$ of the original image $x$ to $x'$,
which yields  
\begin{align}
\tilde x = M(S) \odot x + (1-M(S)) \odot x'.
\label{eq:selective_paste}
\end{align}
Similarly, $M_{ij}(S)=\ind((i,j)\in S)$ is the binary mask of region $S$.

\begin{algorithm}[ht]
\caption{KeepAugment: An information-preserving data augmentation approach}
\label{alg:main}
\begin{algorithmic}
\STATE {\bf Input}: given a network, an input image and label pair $(x, y)$, threshold $\tau$
    \STATE \emph{(a) if use Selective-Cut}
    \STATE ~~~~~~{\bf repeat} randomly select a mask region $S$ {\bf until} region score {$\I(S, x,y)  < \tau$} 
    \STATE  ~~~~~~$\tilde{x} = (1-M(S)) \odot x$ ~(see Eq.~\ref{eq:selective_cut})
    \STATE \emph{(b) if use Selective-paste}
    \STATE ~~~~~~ $x^{\prime} = \mathcal{A}(x)$   ~~~~~~~~~//apply data augmentation
    \STATE ~~~~~~ {\bf repeat} randomly select a mask region $S$ {\bf until} region score {$\I(S, x, y)  > \tau$}
    \STATE ~~~~~~ $\tilde x = M(S) \odot x + (1-M(S)) \odot x'$~~(see Eq.~\ref{eq:selective_paste}) 
\STATE {\bf Return} $\tilde{x}$
\end{algorithmic}
\end{algorithm}

\begin{figure*}[tb]
\centering
\begin{tabular}{ccc}
\raisebox{0.4em}{\rotatebox{90}{Cifar-10 Accuracy}} 
\includegraphics[height=0.2\textwidth]{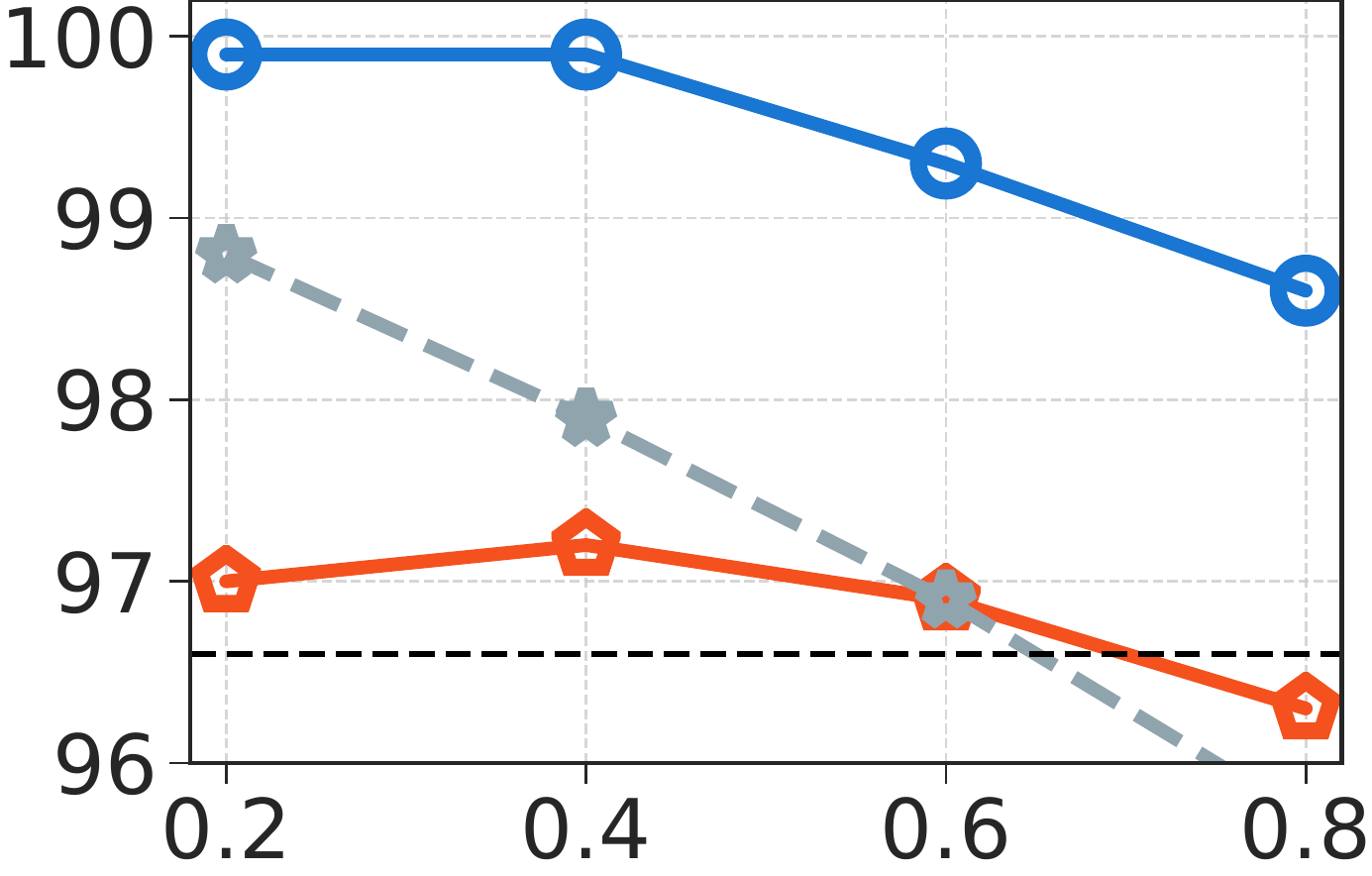}
&
\raisebox{0.4em}{\rotatebox{90}{Cifar-10 Accuracy}} 
\includegraphics[height=0.2\textwidth]{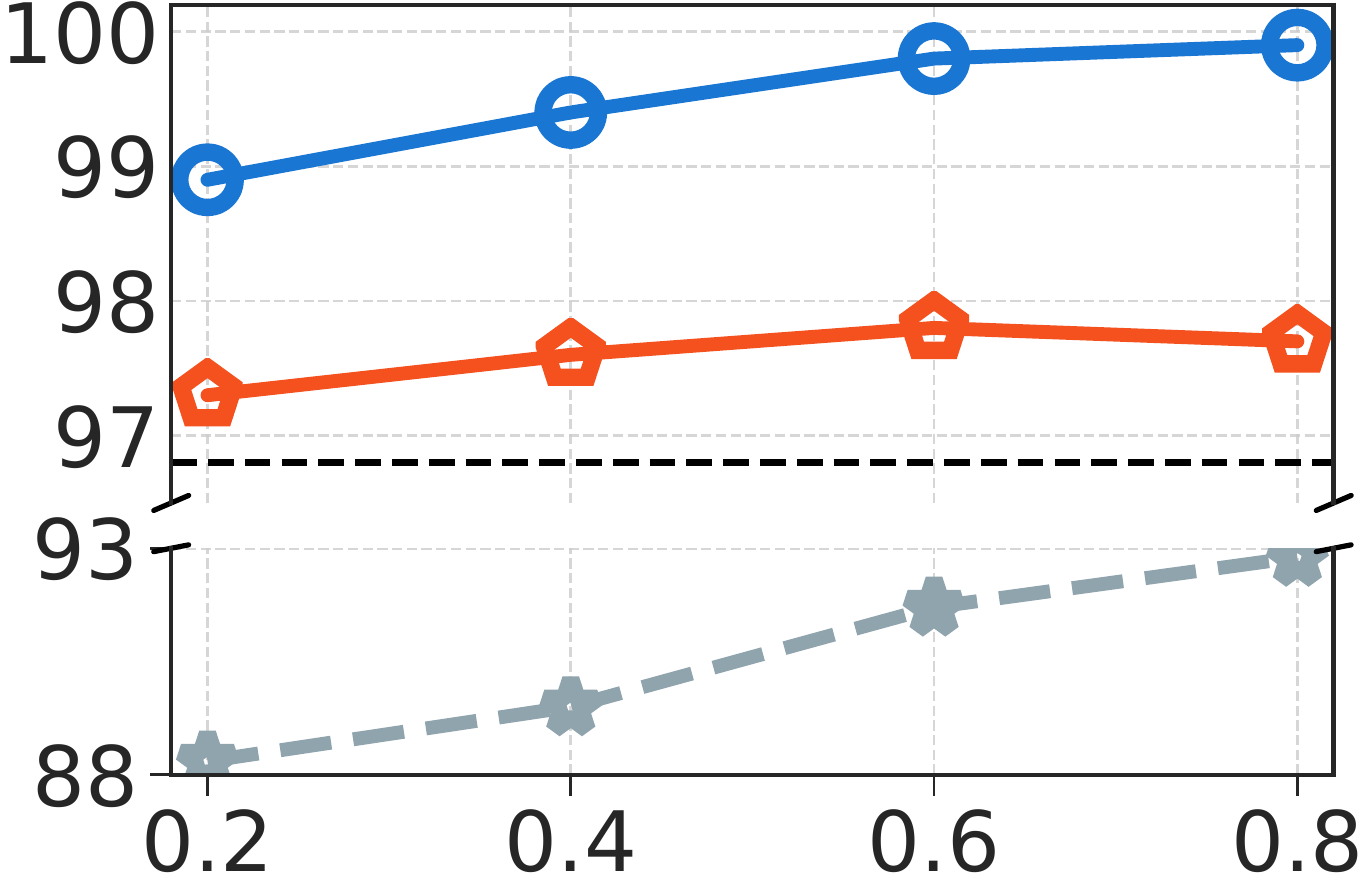}
&
\hspace{-8pt}
\raisebox{1.5em}{ \includegraphics[height=0.1\textwidth]{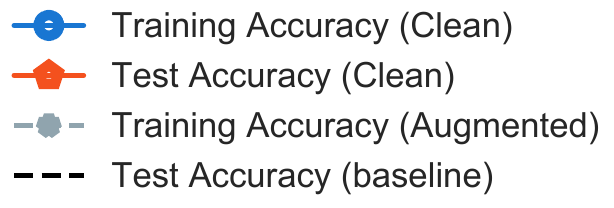}} \\
(a) threshold $\tau$ in CutOut &  (b) threshold $\tau$ in RandAugment \\
\end{tabular}
\caption{Analysis of the effect of threshold $\tau$ of our algorithm for Cutout (a) and  RandAugment (b).  
In (a), we fix the cutout length 20.
In (b), We fix the {number of transformation to be 3 and distortion magnitude to be 15} and the paste back region size to be $8\times 8$. 
We plot how the accuracy changes with respect to different choices of $\tau$.
We use Wide ResNet-28-10 and train on CIFAR-10. 
{The dash line \emph{(baseline)} in (a) represents test accuracy achieved by CutOut without \emph{selective-cut}; the dash line \emph{baseline} in (b) is the test accuracy achieved by RandAugment without \emph{selective-paste}. }
}
\label{fig:study-tau}
\end{figure*}

\paragraph{Remark} 
In practice, we choose our threshold $\tau$ in an adaptive way. 
Technically, given an image and consider an region size $h\times w$ of interest, 
we first calculate the importance scores of all possible candidate regions, following
Eq.~\ref{eq:saliency}; 
then we set our threshold to be the $\tau$-quantile value of 
all the importance scores $\I(S, x, y)$ of all candidate regions.
For \emph{selective-cut}, we uniformly keep sampling a mask region $S$ until its corresponding score $\I(S,x, y)$ is smaller than the threshold. 
For \emph{selective-paste}, 
we uniformly sample a region $S$ with importance score is greater than the threshold.

We empirically study the effect of our threshold $\tau$ on CIFAR-10, illustrated in  Figure \ref{fig:study-tau}. 
Intuitively, for \emph{selective-cut}, 
it's more likely to cut out important regions as we use an increasingly larger threshold $\tau$; on the contrary, a larger $\tau$ corresponds to copy back more critical regions
for  \emph{selective-paste}.
As we can see from Figure~\ref{fig:study-tau}, 
for {\cutout} (Figure~\ref{fig:study-tau} (a)), we improve on the standard {\cutout} baseline (dash line) significantly when the threshold $\tau$ is relative small (e.g. $\tau \le 0.6$) since we would always avoid cutting important regions. 
As expected, the performance drops sharply
when important regions are removed with a relative large threshold $\tau$ (e.g. $\tau=0.8$); 
for {\randaug} (Figure~\ref{fig:study-tau} (b)), using a lower threshold (e.g., $\tau=0.2$) 
tends to yield similar performance as the standard {\randaug} baseline (dash line).
Increasing the threshold $\tau$ ( $\tau$= 0.6 or 0.8)  yields better results. 
We notice that further increasing  $\tau$ ($\tau = 0.8$) may hurt the performance slightly, likely because a large threshold yields too restrictive selection and may miss  other informative regions.

\subsection{Efficient Implementation of KeepAugment}
Note that our KeepAugment requires to calculate the saliency maps via back-propagation at each training step.
Naive implementation leads to roughy twice of the computational cost.
In this part, 
we propose two computational efficient strategies for calculating
saliency maps that overcome this weakness.

\textbf{Low resolution based approximation}
we proceed as follows: a) for a given image $x$, we first generate a low-resolution copy and then calculate its saliency map; b) we map the low-resolution saliency maps to their corresponding original resolution.
This allows us to speed up the saliency maps calculation significantly, e.g., on ImageNet, we achieve 3X computation cost reduction by reducing the resolution from 224 to 112.

\textbf{Early head based approximation}
Our second idea is to introduce an early loss head in the network, 
then we approximate saliency maps with this loss.
In practice, we add an additional average pooling layer and a linear head after the first block of our networks evaluated. 
Our training objective is the same as the Inception Network \cite{szegedy2015going}. The neural network is trained with the standard loss together with the auxiliary loss. We achieve 3X computation cost reduction when calculate saliency map.

Furthermore, 
in the experiment part, we show that both approximation strategies do not lead to any performance drop. 

\vspace{-10pt}
\begin{figure}[h]
\centering
\setlength{\tabcolsep}{1pt}
\begin{tabular}{c}
\includegraphics[width=0.45\textwidth]{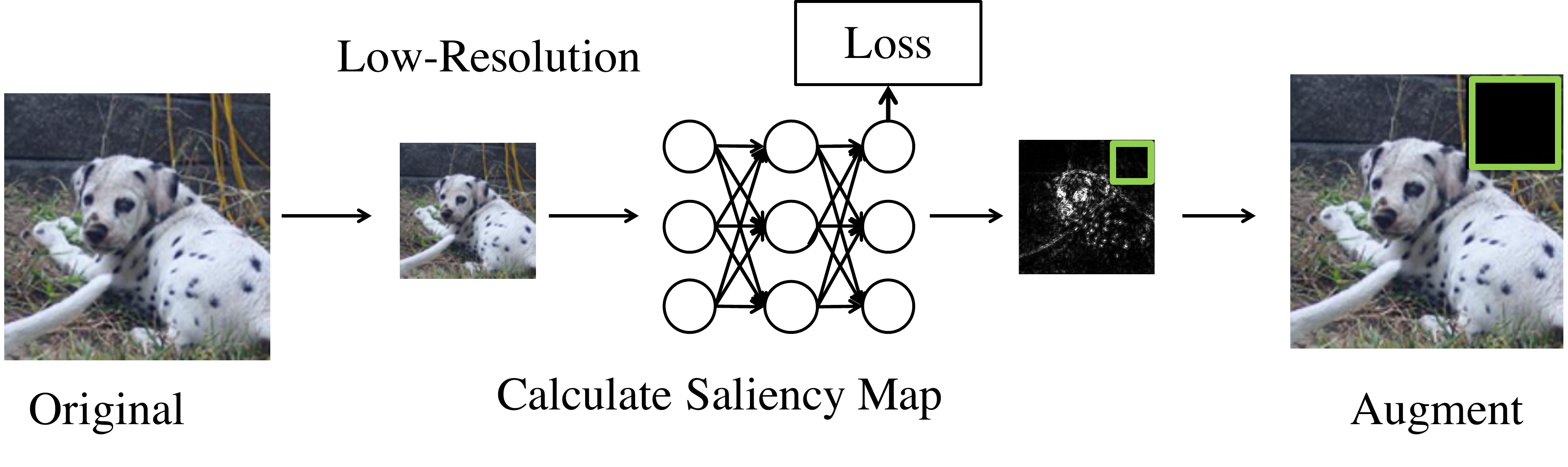} \\
 {\small (a) Low resolution} \\
 \includegraphics[width=0.45\textwidth]{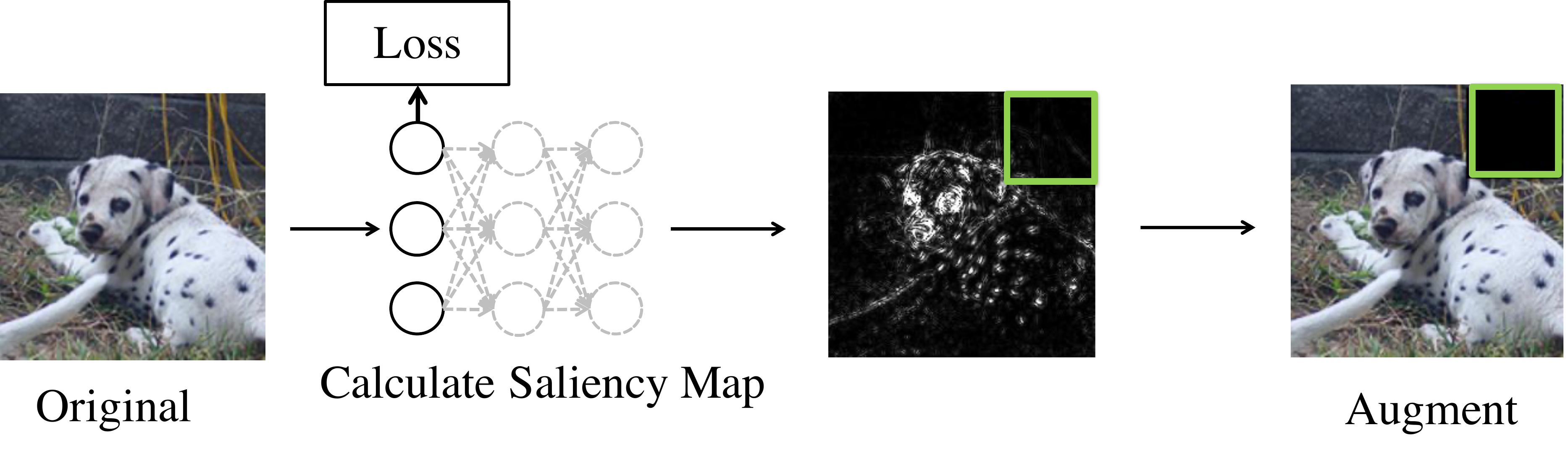} \\
 {\small(b) Early loss}  \\
\end{tabular}
\caption{
We demonstrate two different approaches for using KeepAugment with less training time. Using \cutout as an example, 
Figure (a) shows that we can use a low-resolution copy to calculate the saliency map, and then generate the augmented image.
Figure (b) shows that when calculating the saliency map, we can use an additional loss at early layer of a given neural network.
}
\label{fig:less_time_keepaugmentation}
\end{figure}
\vspace{-5pt}

\begin{table*}[h]
	\begin{center}
    \setlength\tabcolsep{5pt}
	\begin{tabular}{l|ccc}
\hline
	Model & ResNet-18 & ResNet-110 & Wide ResNet-28-10 \\
	\hline
	Cutout & 95.6$\pm$0.1 & 94.8$\pm$0.1 & 96.9$\pm$0.1  \\
	KeepCutout & 96.1$\pm$0.1 & \bf{95.5$\pm$0.1}  & \bf{97.3$\pm$0.1} \\
	KeepCutout  (low resolution) & \bf{96.2$\pm$0.1} & \bf{95.5$\pm$0.1}  & \bf{97.3$\pm$0.1} \\
	KeepCutout  (early loss) & 96.0$\pm$0.1 & 95.3$\pm$0.1  & 97.2$\pm$0.1 \\
\hline \hline
	Model & Wide ResNet-28-10 & Shake-Shake& PyramidNet+ShakeDrop  \\
	\hline
	AutoAugment & 97.3$\pm$0.1 & 97.4$\pm$0.1  & 98.5\\
	KeepAutoAugment & \bf{97.8$\pm$0.1} & 97.8$\pm$0.1 & \bf{98.7$\pm$0.0} \\
	KeepAutoAugment (low resolution) & \bf{97.8$\pm$0.1} & \bf{97.9$\pm$0.1} & \bf{98.7$\pm$0.0} \\
	KeepAutoAugment (early loss) & \bf{97.8$\pm$0.1} & 97.7$\pm$0.1 & 98.6$\pm$0.0 \\
    \hline
	\end{tabular}
	\end{center}
	\caption{Test accuracy (\%) on CIFAR-10 using various models architectures. }
	\label{table:cifar10}  
\end{table*}

\begin{table}[h]
	\begin{center}
	\begin{tabular}{l|cc}
\hline
 Cutout length & Cutout  & KeepCutout\\
 \hline
    8 & \bf{95.3$\pm$0.0} & 95.1$\pm$0.0\\
    12 & 95.4$\pm$0.0 & \bf{95.5$\pm$0.0}\\
    16 & 95.6$\pm$0.0 & \bf{96.1$\pm$0.0}\\
    20 & 95.5$\pm$0.1 & \bf{96.0$\pm$0.1} \\
    24  & 94.9$\pm$0.1 & \bf{95.6$\pm$0.1}\\
    \hline
	\end{tabular}
	\end{center}
	\caption{
	Test accuracy (\%) of ResNet-18 on CIFAR-10 across varying cut-out length. All results are averaged over 5 random trials.}
	\label{table:cifar10-cutout} 
\end{table}

\section{Experiments}

In this section, 
we show our adaptive augmentation strategy \texttt{KeepAugment}  significantly improves on existing state-of-the-art data augmentation baselines 
on a variety of challenging deep learning tasks, including 
image classification, semi-supervised image classification, multi-view multi-camera tracking, and object detection. 
For semi-supervised image classification and multi-view multi-camera tracking, 
we use the low-resolution image to calculate the saliency map as discussed above. 

\paragraph{Settings}

We apply our method to improve 
prior art region-level augmentation methods, including 
\citep{devries2017cutout}, CutMix \citep{yun2019cutmix}, Random Erasing \citep{zhong2017random} and image-level augmentation approach, such as 
AutoAugment \citep{cubuk2018autoaugment}.
To sample the region of interest,
We set $\tau$ to 0.6 for all our experiments (For each image, we rank the absolute saliency values measured on all candidate regions and take our threshold to be the $\tau$-th percentile value.), and 
set the cutout \emph{paste-back} length to be $16$ on CIFAR-10 and $40$ on ImageNet, 
which is the default setting used by \cutout \citep{devries2017cutout}.
For the low-resolution efficient training case, we reduce the image resolution by half with bicubic interpolation.
For the early-loss case, we use an additional head (linear transform and loss) with a coefficient 0.3 after the first block of each network. 



\subsection{CIFAR-10 Classification}

We apply of our adaptive selective strategy to improve two state-of-the-art augmentation schemes, 
{\cutout} and {\randaug}, on the CIFAR-10 \footnote{\url{https://www.cs.toronto.edu/~kriz/cifar.html}} \citep{krizhevsky2010cifar} dataset. 
We experiment with various of backbone architectures, such as ResNets \citep{he2016resnet}, Wide ResNets \citep{zagoruyko2016wide}, PyramidNet ShakeDrop \citep{yamada2018shakedrop} and Shake-Shake \citep{gastaldi2017shake}.
We closely follow the training settings suggested in \citep{devries2017cutout} and \citep{cubuk2019randaugment}.
Specifically, we train 1,800 epochs with cosine learning rate deacy \citep{loshchilov2016sgdr} for PyramidNet-ShakeDrop and 300 epochs for all other networks, 
We report the test test accuracy in Table~\ref{table:cifar10}. 
All results are averaged over three random trials, 
except for PyramidNet-ShakeDrop \citep{yamada2018shakedrop}, on
which only one random trial is reported.

From Table~\ref{table:cifar10}, 
we observe a consistent improvement on test accuracy by applying our information-preserving augmentation strategy. 
In particular, to the best of our knowledge,
our results establish a new state-of-the-art test accuracy $98.7\%$ using 
PyramidNet+Shakedrop among baselines without using extra training data (e.g. ImageNet pretraining). 

\paragraph{Training Time Cost}
Table~\ref{table:cifar10} also displays that, 
the two approaches with less time cost do not hurt the performance on classification accuracy.
The accuracy of the low resolution approach is sometimes even better than the original algorithm.
As shown in Table \ref{table:time}, the original approach (KeepCutout) almost doubles the training cost of \cutout.
By using low-resolution image or early loss head to calculate the saliency map, the training cost is only sightly larger than \cutout, while the performance is improved.

\begin{table}[h]
	\begin{center}
    \setlength\tabcolsep{1pt}
	\begin{tabular}{l|ccc}
\hline
	Model & \small{R-18} & \small{R-110} & \small{Wide ResNet} \\
	\hline
	Cutout & 19 & 28 & 92 \\
	KeepCutout & 38 & 54  & 185 \\
	+ Low Resolution & 24 & 35 & 102 \\
	+ Early Loss & 23 & 34 & 99 \\
    \hline
	\end{tabular}
	\end{center}
	\caption{Per epoch training time (second) on CIFAR-10 using various models architectures. The time (second) is reported on one TITAN X GPU. Here R-18 and R-110 represents ResNet-18 and ResNet-110, respectively.}
	\label{table:time}  
\end{table}

\begin{table}[hbtp]
	\begin{center}
    \setlength\tabcolsep{5pt}
	\begin{tabular}{l|cc}
	\hline
    Magnitude & AutoAugment & KeepAutoAugment
    \\
    \hline
    6 & 96.9$\pm$0.1 & \bf{97.3$\pm$0.1} \\
    12 & 97.1$\pm$0.1 & \bf{97.5$\pm$0.1} \\
    18 & 97.1$\pm$0.1 & \bf{97.6$\pm$0.1} \\
    24 & 97.3$\pm$0.1 & \bf{97.8$\pm$0.1} \\
    \hline
	\end{tabular}
	\end{center}
	\caption{
	Test accuracy (\%) of wide ResNet-28-10 on CIFAR-10 across varying distortion augmentation magnitudes. All results are averaged over 5 random trials.
	}
	\label{table:cifar10-sa} 
\end{table}

\paragraph{Improve on CutOut}
We study the relative improvements on {\cutout} across various cutout lengths. 
We choose ResNet-18 and train on CIFAR-10.
 We experiment with a variety of cutout length from 8 to 24.
 As shown in Table ~\ref{table:cifar10-cutout}, 
 we observe that our KeepCutout achieves increasingly significant improvements
 over {\cutout} when the cutout regions become larger. 
This is likely because that with large cutout length, 
{\cutout} is more likely to remove the most informative region
and hence introducing misinformation, which in turn hurts the network performance. 
On the other hand, with a small cutout length, e.g. 8, 
those informative regions are likely to be preserved during augmentation; 
standard {\cutout} strategy achieves better performance by taking advantage of
more diversified training examples. 

\vspace{-5pt}
\paragraph{Improve on AutoAugment}
In this case, we use the AutoAugment policy space, apply our \emph{selective-paste} and study the empirical gain over AutoAugment for four distortion 
augmentation magnitude (6, 12, 18 and 24). 
We train Wide ResNet-28-10 on CIFAR-10 and closely follow the training setting suggested in \citep{cubuk2018autoaugment}. 
As we can see from Table~\ref{table:cifar10-sa}, 
our method yields better performance in all settings consistently, 
and our improvements is more significant when the transformation distortion  
magnitude is large.

\begin{table}[h]
	\begin{center}
    \setlength\tabcolsep{5pt}
	\begin{tabular}{l|ccc}
    \hline
	Wide ResNet-28-10 & Accuracy (\%) & Time (s)\\
	\hline
	GridMask & 97.5$\pm$0.1 & 92\\
	AugMix & 97.5$\pm$0.0 & 92 \\
	Attentive CutMix & 97.3$\pm$0.1 & 127  \\
	KeepAutoAugment+L & \bf{97.8$\pm$0.1} & 102 \\
    \hline
	ShakeShake & Accuracy (\%) & Time (s)\\
	\hline
	GridMask & 97.4$\pm$0.1 & 124\\
	AugMix & 97.5$\pm$0.0 & 124 \\
	Attentive CutMix & 97.4$\pm$0.1 & 166  \\
	KeepAutoAugment+L & \bf{97.9$\pm$0.1} & 142 \\
    \hline
	\end{tabular}
	\end{center}
	\vspace{5pt}
	\caption{Results on CIFAR-10 using various models architectures and  various baselines. `Time' reports the per epoch training time on one TITAN X GPU. `Accuracy' reports the accuracy on test set, which is averaged over 5 trials. `L' denotes low resolution.}
	\label{table:cifar10-baselines}  
\end{table}

\vspace{-5pt}
\paragraph{Additional Comparisons on CIFAR-10}

Recently, some researchers \citep{chen2020gridmask, walawalkar2020attentive, hendrycks2019augmix} also mix the clean image and augmented image together to achieve higher performance. 
Girdmask \citep{chen2020gridmask}, AugMix \citep{hendrycks2019augmix} and Attentive \cutmix are  popular methods among these approaches.
(Please refer to Section \ref{sec:related} for more details about \cutmix.)
Here, we do experiments on CIFAR-10 to show the accuracy and training cost of each method.
Note that 
we implement all the baselines by ourselves, 
and the results of our implementation are comparable or even better 
than the results reported in the original papers.

Table \ref{table:cifar10-baselines} shows that the proposed algorithm can achieve clear improvement on accuracy over all the baselines. 
Using the low resolution image to calculate the saliency map, 
our method does not introduce additional time cost to the training process.
On the other hand, 
while Gridmask only implements upon \cutout~ and 
Attentive \cutmix~ only implements upon \cutmix~ by pasting the most important region, 
our approach is more flexible and can be applied to other data augmentation approaches.

\subsection{ImageNet Classification}
\label{sec:imagenet}
We conduct experiments on large-scale challenging ImageNet dataset, on which our 
adaptive augmentation algorithm again shows clear advantage over existing methods.

\begin{table}[h]
    \centering
    \begin{tabular}{l|cc|cc}
        \hline
        \multirow{2}{*}{Method} & \multicolumn{2}{c|}{ResNet-50} &  \multicolumn{2}{c}{ResNet-101}\\
        \cline{2-5}
        & Top-1 & Top-5 & Top-1  & Top-5  \\
        \hline
        Vanilla {~\citep{he2016resnet}} & 76.3 & 92.9 & 77.4 & 93.6 \\
        Dropout {~\citep{srivastava2014dropout}} & 76.8 & 93.4 & 77.7 & 93.9 \\
        DropPath {~\citep{larsson2016fractalnet}} & 77.1 & 93.5 & - & -\\
        Manifold Mixup {~\citep{verma2018manifold}} & 77.5 & 93.8 & - & - \\
        Mixup {~\citep{zhang2018mixup}} & 77.9 & 93.9 & 79.2 & 94.4\\
        DropBlock {~\citep{ghiasi2018dropblock}} & 78.3 & 94.1 & 79.0 & 94.3\\
        RandAugment ~\citep{cubuk2019randaugment} & 77.6 & 93.8 & 79.2 & 94.4\\
        \hline
        AutoAugment {~\citep{cubuk2018autoaugment}} & 77.6 & 93.8 & 79.3 & 94.4 \\
        KeepAutoAugment & 78.0 & \bf{93.9} & \bf{79.7} & \bf{94.6} \\
        + Low Resolution & \bf{78.1} & \bf{93.9} & \bf{79.7} & \bf{94.6} \\
        + Early Loss & 77.9 & 93.8 & 79.6 & 94.5 \\
        \hline
        CutMix {~\citep{yun2019cutmix}} & 78.6 & 94.0 & 79.9 & 94.6 \\
        KeepCutMix & 79.0 & \bf{94.4} & \bf{80.3} & 95.1 \\ 
        + Low Resolution & \bf{79.1} & \bf{94.4} & \bf{80.3} & \bf{95.2} \\
        + Early Loss & \bf{79.0} & 94.3 & 80.2 & 95.1 \\
        \hline
    \end{tabular}
    \caption{Validation set Top-1 and Top-5 accuracy (\%) on ImageNet using ResNet-50 and ResNet-101.
    }
    \label{tab:imagenet-resnet}
\end{table}

\paragraph{Dataset and Settings}
We use  ILSVRC2012, a subset of ImageNet classification dataset \citep{deng2009imagenet}, which
contains around 1.28 million training images and 50,000 validation images from 1,000 classes. 
We apply our adaptive data augmentation strategy to improve {\cutmix} \citep{yun2019cutmix} and {\autoaug} \citep{cubuk2018autoaugment}, respectively. 

{\cutmix} randomly mixes images and labels. 
To augment an image $x$ with label $y$,
{\cutmix}  removes a randomly selected region from $x$ and 
replace it with a patch of the same size 
copied from another random image $x'$ with label $y'$. 
Meanwhile, the new label is mixed as $\lambda y + (1-\lambda)y'$, where 
$\lambda$ equals the uncorrupted percentage of image $x$.
We improve on {\cutmix} by using \emph{selective-cut}. 
In practice, we found it is often quite effective to simply avoiding cutting informative region from $x$, as we observe the mixing rate $\lambda$ sampled is often close to $1$.
We denote our adaptive CutMix method as \texttt{KeepCutMix}.
We further improve on {\autoaug} by pasting-backing randomly selected regions with important score greater than $\tau = 0.6$. 

For a fair comparison, we closely follow the training settings in \texttt{CutMix} \citep{yun2019cutmix} and {\autoaug} \citep{cubuk2019randaugment}. 
We test our method on both ResNet50 and ResNet101 \citep{he2016resnet}.
Our models are trained for 300 epochs, and the experiment is implemented based on the open-source code \footnote{\url{https://github.com/clovaai/CutMix-PyTorch}}.

\paragraph{Results} 
We report the single-crop top-1 and top-5 accuracy on the validation set in table~\ref{tab:imagenet-resnet}.
Compared to \texttt{CutMix}, we method \texttt{KeepCutMix} achieves 0.5\% improvements on top1 accuracy using ResNet-50 and 0.4\% higher top1 accuracy using ResNet101; compared to AutoAugment \citep{cubuk2018autoaugment}, our method improves top-1 accuracy from 77.6\% to 78.1\% and 79.3\% to 79.7\% using ResNet-50 and ResNet-101, respectively.
Again, we also notice that our accelerated approaches do not hurt the performance of the model. We also notice that, similar to the results on CIFAR-10, the proposed accelerating approach can speed up KeepAugment without loss of accuracy on ImageNet.


\subsection{Semi-Supervised Learning}
Semi-supervised learning (SSL) is a key approach toward more data-efficient machine learning by jointly leverage both labeled and unlabeled data. 
Recently, data augmentation has been shown a powerful tool for developing  state-of-the-art SSL methods.
Here, 
we apply the proposed method to unsupervised data augmentation \citep{xie2019uda} (UDA) on CIFAR-10 to verify whether our approach can be applied to more general applications.

UDA minimizes the following loss on unlabelled data:
$
\mathbb{E}_{ x \sim \mathcal{D}_u, ~ x' \sim \mathcal{P}_x} \bigg [KL(p_{\theta}(\cdot~|~x) ~\parallel~ p_\theta(\cdot ~|~ x')) \bigg]
$, 
where $\mathcal{P}$ denotes the randomized augmentation distribution, $x'$ denotes an augmented image and $\theta$ denotes the neural network parameters.
Notice that for semi-supervised learning, we do not have labels to calculate the saliency map. Instead, we use the max logit of $p_\theta(\cdot ~|~x)$ to calculate the saliency map.
We simply replace the RandAugment \citep{cubuk2019randaugment} in UDA with our proposed approach, and use the WideResNet-28-2.

Table \ref{tab:semi} shows that our approach can consistently improve the UDA with different number of labelled image on CIFAR-10.

\begin{table}[h]
    \centering
    \begin{tabular}{l|cc}
        \hline
        & 4000 labels & 2500 labels \\
        \hline
        UDA + RandAug & 95.1$\pm$0.2 & 91.2$\pm$1.0\\
        UDA + KeepRandAug & \bf{95.4$\pm$0.2} & \bf{92.4$\pm$0.8} \\
        \hline
    \end{tabular}
    \caption{Result on CIFAR-10 semi-supervised learning. `4000 labels' denotes that 4,000 images have labels while the other 4,6000 do not.}
\label{tab:semi}
\end{table}


\subsection{Multi-View Multi-Camera Tracking}
We apply our adaptive data augmentation to improve a state-of-the-art multi-view multi-camera tracking approach \citep{luo2019bag}.
Recent works \citep[e.g.][]{luo2019bag, zhong2017random, zheng2015market1501} have shown that data augmentation is an effective technique for improving the performance on this task. 

\paragraph{Settings}~~
\citep{luo2019bag} builds a strong baseline based on \texttt{Random Erasing} \citep{zhong2017random} data augmentation.  \texttt{Random Erasing} is similar to {\cutout}, except filling the region dropped with random values instead of zeros.
We improve over \citep{luo2019bag} by only cutting out regions with importance score smaller than $\tau = 0.6$. 
We denote the widely-used open-source baseline \emph{open-ReID} \footnote{\url{https://github.com/Cysu/open-reid}} as the standard baseline
in table ~\ref{tab:person_reid}. 
To ablate the role of our selective cutting-out strategy, we pursue minima changes made to the baseline code base. We follow all the training settings reported in \citep{luo2019bag}, 
except using our adaptive data augmentation strategy. 
We use ResNet-101 as the backbone network.

\begin{table}[h]
    \centering
    \begin{tabular}{l|cc}
        \hline
        \multirow{2}{*}{Method} & \multicolumn{2}{c}{Market1501}\\
        \cline{2-3}
        & Accuracy & mAP \\
        \hline
        Standard Baseline  & 88.1$\pm$0.2 & 74.6$\pm$0.2\\
        + Bag of Tricks \citep{luo2019bag} & 94.5$\pm$0.1 & 87.1$\pm$0.0 \\
        \hline
        + Ours & \bf{95.0$\pm$0.1} & \bf{87.4$\pm$0.0} \\ 
        \hline
    \end{tabular}
    \caption{We compare our method with the standard and \citep{luo2019bag} on two benchmark datasets. \emph{mAP} represents mean average precision.}
    \vspace{-10pt}
\label{tab:person_reid}
\end{table}\begin{table*}[h]
    \centering
    \begin{tabular}{l|c|cc}
    \hline
    Model & Backbone & Detectron2 (mAP\%) & Ours (mAP\%) \\ \hline
    Faster R-CNN & ResNet50-C4 &  38.4 & \bf{39.5}  \\
    Faster R-CNN & ResNet50-FPN & 40.2 & \bf{40.7} \\
    RetinaNet & ResNet50-FPN & 37.9 & \bf{39.1} \\
    \hline
    Faster R-CNN & ResNet101-C4 & 41.1 &  \bf{42.2} \\
    Faster R-CNN & ResNet101-FPN & 42.0 & \bf{42.9} \\
    RetinaNet & ResNet101-FPN & 39.9 & \bf{41.2} \\
    \hline
    \end{tabular}
    \caption{ Detection mean Average Precision (mAP) Results on COCO 2017 validation set. 
    }
    \label{tab:object_detection}
    \vspace{-10pt}
\end{table*}
\paragraph{Dataset}~~
We evaluate our method
on a widely used benchmark dataset, Market1501 \citep{zheng2015market1501}.
Market1501 contains 32,668 bounding boxes of 1,501 identities, in which images of each identity are captured by at most six cameras. 

\paragraph{Results}~~
We report the accuracy and 
mean average precision (mAP) of different methods in  Table \ref{tab:person_reid}. 
Our method achieves the best performance on both datasets. 
In particular, 
we achieve a 95.0\% accuracy and 87.4 mAP on Market1501. 

\subsection{Transfer Learning: Object Detection}
We demonstrate the transferability of our ImageNet pre-trained models on the COCO 2017 \citep{lin2014microsoft} object detection task, on which we observe significant improvements
over strong \texttt{Detectron2} \citep{wu2019detectron2} baselines by simply 
applying our pre-trained models as backbones. 

\paragraph{Dataset and Settings}~~
We use the COCO 2017 challenge, which consists of 118,000 training images and 5,000 validation images. 
To verify that our trained models can be widely useful for different detector systems, 
we test several popular structures, including Faster RCNN \citep{ren2015faster}, feature pyramid networks \citep{lin2017feature} (FPN) and RetinaNet \citep{lin2017focal}.
We use the codebase provided by \texttt{Detectron2} \citep{wu2019detectron2}, follow almost all the hyper-parameters except changing the backbone networks from \texttt{PyTorch} provided models to our models. For our method, we test the ResNet-50 and ResNet-101 models trained with our \texttt{KeepCutMix}.

\paragraph{Results}~~
We report mean average precision (mAP) on the COCO 2017 validation set
\citep{lin2014microsoft}.
As we can see from Table~\ref{tab:object_detection}, our method consistently improves over baseline approaches.
%
Simply replacing the backbone network with our pre-trained model gives performance gains for the COCO 2017 object detection tasks with no additional cost. 
In particular, on the single-stage detector RetinaNet, we improve the 37.9 mAP to 39.1, and 39.9 mAP to 41.2 for ResNet-50 and ResNet-101, respectively,
while the pre-trained model processed by \cutmix~ cannot even match the performance of the \texttt{Detectron2} baselines.
ResNet-50 + \cutmix achieves 23.7, 27.1 and 25.4 (mAP) in our C4, FPN and RetinaNet settings, respectively.  These results are much worse compared to our results. 

\section{Related Works}
\label{sec:related}

Our work is most related to \citep{gontijo2020affinity}, 
which studies the impact of affinity (or fidelity) and
diversity of data augmentation empirically, 
and finds out that a good data augmentation strategy should 
jointly optimize these two aspects. 
Recently, many other works also show the importance of balancing between fidelity and diversity.
For example, \citep{gong2020maxup} and \citep{zhang2019adversarial} show that optimize the worst case or choose the most difficult augmentation policy is helpful, which indicates the importance of diversity.

\citep{walawalkar2020attentive} also considers to extract the most informative region to reduce noise in {\cutmix}. 
Their proposed method, Attentive \cutmix requires an additional pretrained classification model to serve as a teacher, while our approach does not need additional supervision. 
Moreover, 
our approach is much more general and can be applied to more augmentation methods and tasks (e.g. semi-supervision) which Attentive \cutmix cannot. 
\citep{wei2020circumventing} considers to correct the \emph{label} of noisy augmented examples by using a teacher network, thus increasing fidelity. 
This approach also needs additional supervision and only focus on one typical data augmentation method.
Compared to these works, 
our augmentation improves on stronger data augmentation by preserving informative regions, thus 
naturally achieve fidelity and diversity. 
It allows us to train better models by leveraging more diversified faithful examples. 

Our work focus on improving label-invariant data augmentation. Another line of data augmentation schemes create augmented examples by mixing both images and their corresponding labels, exemplified by \emph{mixup} \citep{zhang2018mixup}, 
Manifold Mixup \citep{verma2018manifold}, CutMix \citep{yun2019cutmix}. 
It is not clear how to quantify noisy examples for label-mixing augmentation since labels
are also mixed, nevertheless we show empirically that our \emph{selective-cut} also improves on {\cutmix} and leave further extensions as our future work. 

The idea of using saliency map for improving computer vision systems have been widely explored in the literature. 
Saliency map can be applied to object detection \citep{zhang2013saliencydetect}, segmentation \citep{mukherjee2015saliencyseg}, knowledge distillation \citep{chan2019thinks} and many more \citep[e.g.][]{chan2019thinks, selvaraju2017grad}.
We propose to use the saliency map to to measure the relative importance of different regions  of the input, thus improving regional-level cutting-based data augmentation 
by avoiding informative regions;  or improving image-level augmentation techniques by 
pasting-back discriminative regions.


\section{Conclusion}
In this work, 
we empirically show that prior art data augmentation schemes might introduce noisy training examples and hence limit their ability in boosting the overall performance. 
Thus we use saliency map to measure the importance of each region, 
and propose to: avoid cutting important regions for region-level data augmentation approaches, such as {\cutout}; or pasting back critical areas from the clean data for image-level data augmentation, like {\randaug} and {\autoaug}. 
Throughout an extensive evaluation, we have demonstrated that our adaptive 
augmentation approach helps to significantly improve the performance of state-of-the-art architectures on CIFAR-10, ImageNet, multi-view multi-camera tracking and object detection. 

\newpage
{\small
\bibliographystyle{ieee_fullname}
\bibliography{egbib}
}

\end{document}